\documentclass[sigconf]{acmart}
\newcommand{\bbf}[1]{\underline{#1}}
\usepackage{multirow}
\usepackage{subfig}
\usepackage{graphicx}
\usepackage{algorithmic}
\usepackage{algorithm}
\usepackage{hyperref}

\newlength\savewidth\newcommand\shline{\noalign{\global\savewidth\arrayrulewidth
  \global\arrayrulewidth 1pt}\hline\noalign{\global\arrayrulewidth\savewidth}}
\newcommand{\tablestyle}[2]{\setlength{\tabcolsep}{#1}\renewcommand{\arraystretch}{#2}\centering\footnotesize}

\AtBeginDocument{%
  }

\copyrightyear{2024}
\acmYear{2024}
\setcopyright{acmlicensed}\acmConference[SIGIR '24]{Proceedings of the 47th International ACM SIGIR Conference on Research and Development in Information Retrieval}{July 14--18, 2024}{Washington, DC, USA}
\acmBooktitle{Proceedings of the 47th International ACM SIGIR Conference on Research and Development in Information Retrieval (SIGIR '24), July 14--18, 2024, Washington, DC, USA}
\acmDOI{10.1145/3626772.3657823}
\acmISBN{979-8-4007-0431-4/24/07}

\begin{document}

\title{CaLa: Complementary Association Learning for Augmenting Composed Image Retrieval
}


\author{Xintong Jiang \footnotemark[1]} 
\email{jiangxintong@stu.xjtu.edu.cn}
\orcid{0009-0002-8664-460X}
\affiliation{%
  \institution{Xi'an Jiaotong University}
  \streetaddress{No.28 Xianning West Road}
  \city{Xi'an}
  \state{Shaanxi}
  \country{P.R. China}
  \postcode{710049}
}

\author{Yaxiong Wang \footnotemark[1]} 
\email{wangyx15@stu.xjtu.edu.cn}
\affiliation{%
  \institution{Hefei University of Technology}
  \streetaddress{No. 193 Tunxi Road}
  \city{Hefei}
  \state{Anhui}
  \country{P.R. China}
  }

\author{Mengjian Li}
\email{limengjian@zhejianglab.com}
\affiliation{%
  \institution{Zhejiang Lab}
  \streetaddress{Kechuang Avenue, Zhongtai}
  \city{Hangzhou}
  \state{Zhejiang}
  \country{P.R. China}
  }
  
\author{Yujiao Wu \footnotemark[2]}
\email{yujiaowu111@gmail.com
}
\affiliation{%
  \institution{CSRIO}
  \city{Hobart}
  \country{Australia}
  }

\author{Bingwen Hu}
\email{hubw.sky@gmail.com
}
\affiliation{%
  \institution{Anhui University of Technology}
  \streetaddress{No. 193 Tunxi Road}
  \city{Ma'anShan}
  \state{Anhui}
  \country{P.R. China}
  }

\author{Xueming Qian \footnotemark[2]}
\email{qianxm@mail.xjtu.edu.cn}
\affiliation{%
  \institution{Xi'an Jiaotong University}
  \streetaddress{No.28 Xianning West Road}
  \city{Xi'an}
  \state{Shaanxi}
  \country{P.R. China}
  \postcode{710049}
  }


\renewcommand{\shortauthors}{Xintong Jiang et al.}

\begin{abstract}
Composed image retrieval (CIR) is the task of searching target images using an image-text pair as a query. Given the straightforward relation of query pair-target image, the dominant methods follow the learning paradigm of common image-text retrieval and simply model this problem as the query-target matching problem. 
Particularly, the common practice first encodes the multi-modal query into one feature and then aligns it with the target image. However, such a learning paradigm only explores the na\"ive relation in the triplets.
We argue that CIR triplets encompass additional associations besides the primary query-target relation, which is overlooked in existing works.
In this paper, we disclose two new relations residing in the triplets by viewing the triplet as a graph node. In analogy with the graph node, we mine two associations of text-bridged image alignment and complementary text reasoning. 
The text-bridged image alignment considers composed image retrieval as a specialized form of image retrieval, where the query text acts as a bridge between the query image and the target one, and a hinge-based cross attention is proposed to incorporate this relation into the network learning. On the other hand, the association of complementary text reasoning regards composed image retrieval as a specific type of cross-modal retrieval, where the composite two images are used to reason the complementary text. To integrate these views effectively, a twin attention-based compositor is designed. By combining these two types of complementary associations with the explicit query pair-target image relation, we establish a comprehensive set of constraints for composed image retrieval. With the above designs, we finally developed our CaLa, a Complementary Association Learning framework for Augmenting Composed Image Retrieval.
Experimental evaluations are conducted on the widely-used CIRR and FashIionIQ benchmarks with multiple backbones to validate the effectiveness of our CaLa. The results demonstrate the superiority of our method in the composed image retrieval task. Our code and models are available at 
\href{https://github.com/Chiangsonw/CaLa}{https://github.com/Chiangsonw/CaLa}

\end{abstract}


\maketitle

\renewcommand{\thefootnote}{\fnsymbol{footnote}}
\footnotetext[1]{Equal contribution}
\footnotetext[2]{Corresponding author}
\renewcommand*{\thefootnote}{\arabic{footnote}}

\begin{figure}[t] 
\centering 
\includegraphics[width=0.44 \textwidth]{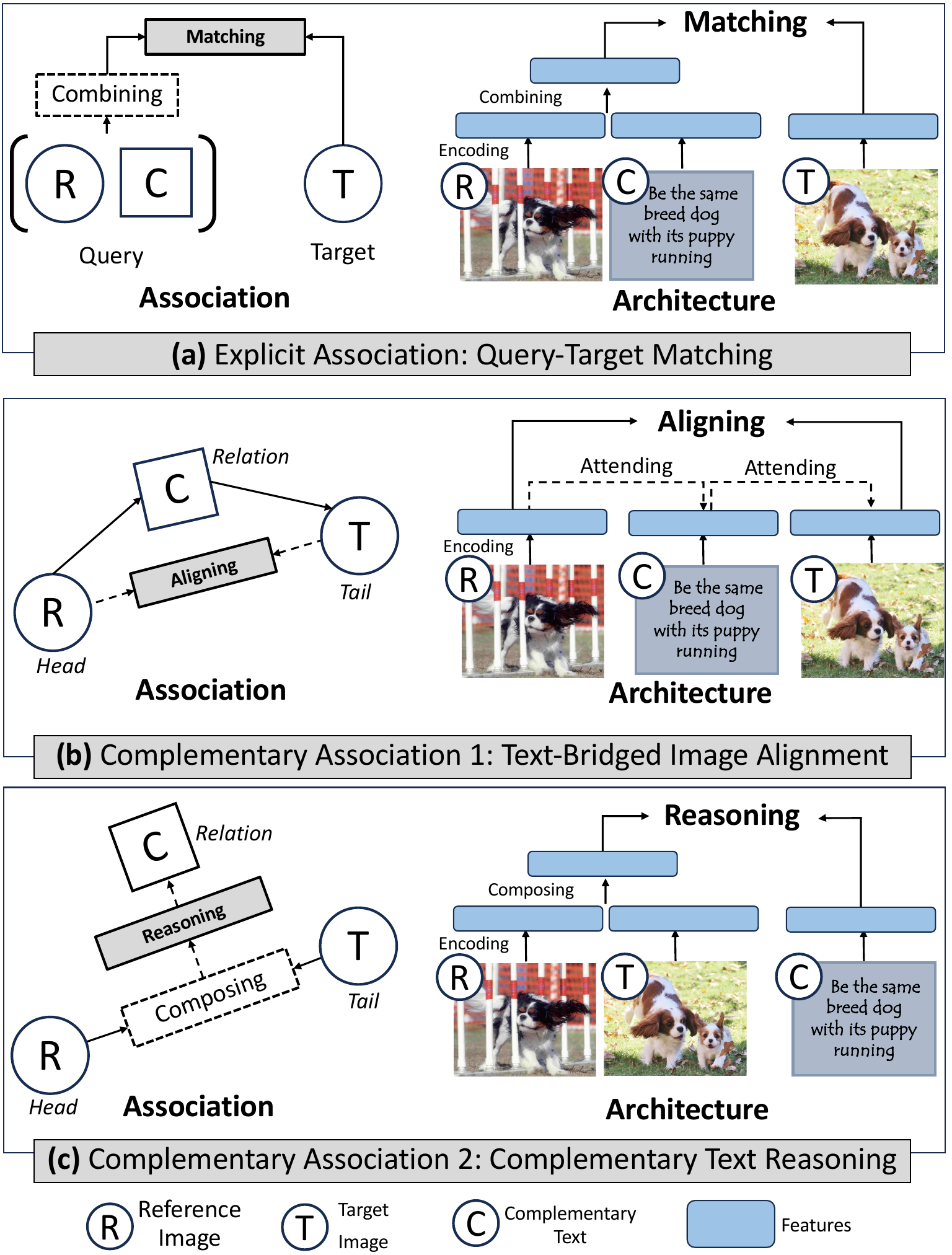} 
\caption{The illustration of the popular explicit association (a) and our disclosed two complementary associations in this paper (b)-(c), where the left part and right part separately show the association and architecture for integration of the constraints. The explicit association in (a) is typically modeled as the query-target matching problem. By considering the triplet as a graph node, we disclose two new associations of text-bridged image alignment (b) and complementary text reasoning (c) and integrate them into network learning via a hinge-based cross-attention and twin attention-based compositor.
} 
\label{motivation} 
\end{figure}

\section{Introduction}

Composed Image Retrieval (CIR) is a subfield of cross-modal retrieval~\cite{pfan,pfan1}, in which the query comprises a reference image and complementary text, providing richer context and additional information for the retrieval process.
Compared to single-modality retrieval, CIR leverages information from both reference image and text inputs, facilitating more fine-grained and precise retrieval. In recent years, CIR has garnered significant attention due to its potential to disrupt traditional retrieval scenarios~\cite{liu2021image,baldrati2022conditioned,delmas2022artemis,lee2021cosmo,zhang2023relieving}, such as search engines and e-commerce platforms, offering users a novel means to articulate their preferences on websites.


CIR is closely related to the well-studied image-text matching (ITM) problem~\cite{zzd,reidyaxiong1,shuyu,qu2023learnable}, consequently, many existing solutions for CIR adopt learning paradigms from ITM. For instance, the notable method, CLIP4Cir~\cite{baldrati2021conditioned,baldrati2022conditioned},  encodes the multimodal query as a single representation and aligns it with the target image via contrastive learning, akin to conventional image-text matching. However, these methods primarily focus on the explicit query-target relation, neglecting the richer associations within the CIR triplet. While CIR and ITM share similarities, the triplet in CIR encompasses more intricate relations beyond the na\"ive image-text pair in ITM. Therefore, treating CIR as a query-target matching task and solely utilizing the explicit relation is the one-side solution.  However, existing works spare rare efforts to mine the complementary association in the triplet, resulting in suboptimal learning strategies.

In contrast to the tightly bound image-text pair in ITM, the triplet in CIR is more analogous to a graph node, with the reference image (head) connected to the target image (tail) via the complementary text (relation), as shown in Figure\ref{motivation} (b). This perspective suggests that CIR can be regarded as a special case of the image retrieval problem, where the complementary text plays a crucial role as a bridge for similarity estimation.  From this view, we propose to align the reference image and the target image by leveraging the guidance of the complementary text. To impose this constraint, we design a hinge-based cross-attention (HCA). Specifically, hinge-based cross-attention first computes the attention between the reference and the target images using the complementary text as a pivot point. Then, the attention matrix is utilized to highlight the patches of the target image, which is then aligned with the reference image. Finally, a contrastive procedure is employed between the reference and target images to integrate the constraint.

In a graph node (head, relation, tail), the combination of head and tail can infer the relation, this is a commonly-used clue in graph learning community~\cite{graph1,graph2,graph3}. We found this type of association also exists in the triplets of the CIR problem.  As depicted in Figure~\ref{motivation} (a), the multimodal query can locate the target image, which is the widely-used explicit relation in CIR.
On the other hand, by comparing the two images, we can also deduce that if we want to locate the target using the reference image, the modification text should express something like “Be a same breed dog with its puppy running” (Figure~\ref{motivation} (c)). This implies the presence of a potential relation between the images and the complementary text, which, unfortunately, has been largely overlooked in the composed image retrieval community. To impose this type of association,  we design a twin attention-based vision compositor, which effectively merges the reference and target images. Subsequently, the resulting representation serves as the visual counterpart to the complementary text. Finally, we perform alignment learning between the combined visual feature and the complementary text, with a focus on the association revealed by this implicit relation.

Taking the above designs and constraints into the learning of composed image retrieval, we finally develop \textbf{CaLa}, a \textbf{C}omplementary \textbf{A}ssociation \textbf{L}earning framework for \textbf{A}ugmenting the composed image retrieval.  In summary, we highlight the contributions of this paper as follows:
\begin{itemize}
    \item We present a new thinking of composed image retrieval, the annotated triplet is viewed as a graph node, and two complementary association clues are disclosed to enhance the composed image retrieval.

    \item A hinge-based attention and twin-attention-based visual compositor are proposed to effectively impose the new associations into the network learning.
    
    \item Competitive Performance on CIRR and FashionIQ benchmarks. CaLa can benefit several baselines with different backbones and architectures, revealing it is a widely beneficial module for composed image retrieval. 
    
\end{itemize}

\begin{figure*}[t] 
\centering 
\includegraphics[width= 0.9\textwidth]{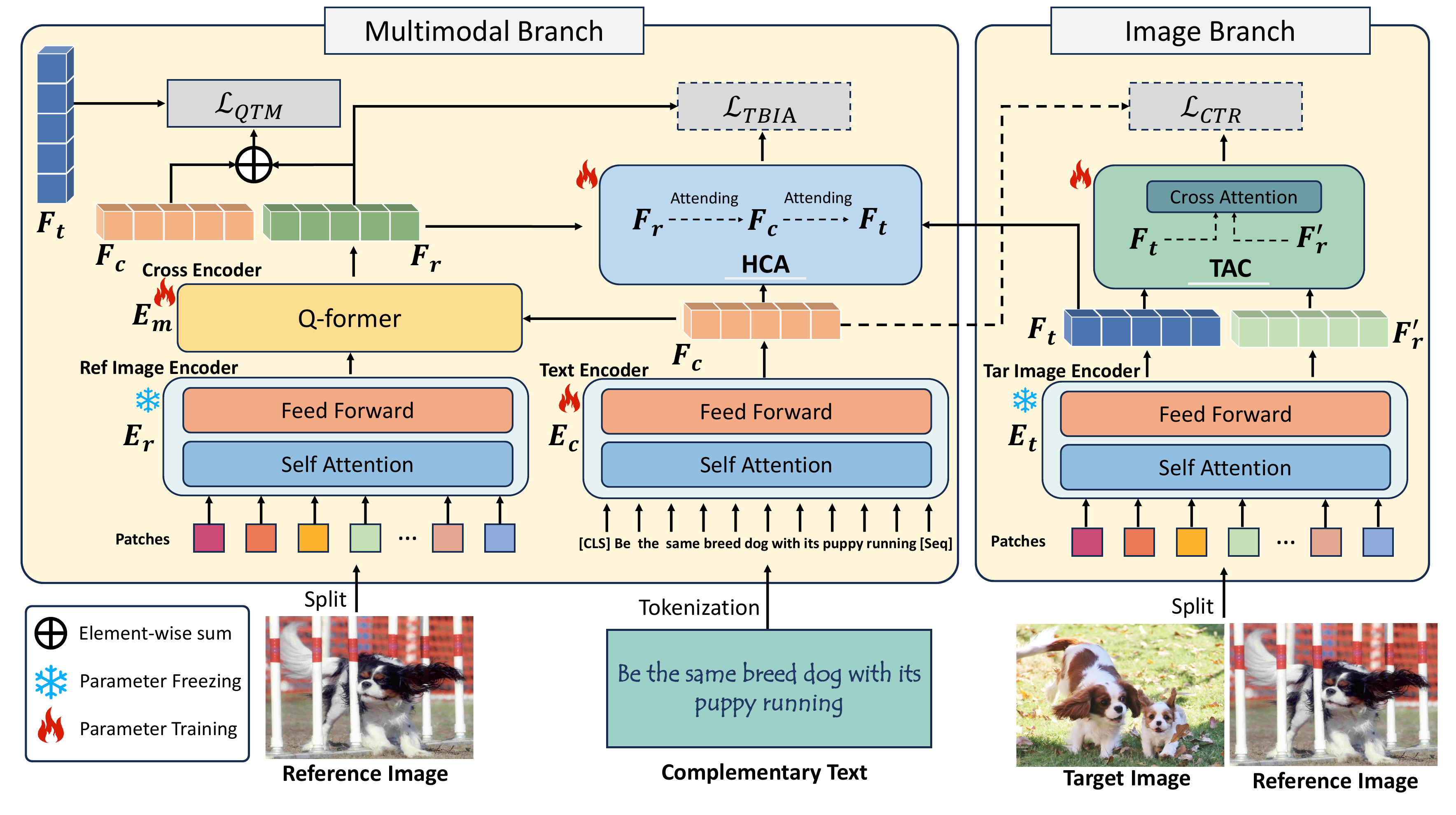} 
\vspace{-0.4cm}
\caption{\textbf{Illustration of our CaLa architecture.} CaLa is a two-branch architecture, where a multimodal branch and image branch serve for the query and target image feature extraction, respectively. Given a query pair and the matched target image, their features are first extracted with respective encoders. With these representations, the proposed hinge-based cross-attention (HCA) module and twin attention-based vision compositor (TAC) module are equipped on the top of the base encoders, imposing the two complementary associations. Note that the data flows for our complementary association integration are only applied in the traning stage (dashed boxes), introducing no inference burden. 
} 
\label{flowchart} 
\end{figure*}

\section{Related Works}
{Composed image retrieval has garnered substantial attention in recent years, and significant efforts have been devoted to advancing this field. 
For example, CLIP4Cir~\cite{baldrati2022conditioned} employs CLIP~\cite{clip} as the underlying network and trains an image-text combiner to integrate the reference image and complementary text in a unified representation, which is then matched with the target image.
\\Recent works in composed image retrieval have benefited from advancements in cross-modal pretraining~\cite{clip,blip,coca,wang2023image, wang2023all}.
For instance, BLIP4CIR+Bi ~\cite{liu2024bi} exploits information in the mapping from the ⟨target
image, reversed modification text⟩-pair to the reference image, which provides a new perspective as a complementary relation.

Works like \cite{wen2023target, chen2024spirit} are also exploring hidden relations in the image-text tuples in unique perspectives.
CASE\cite{levy2023data} also leveraged the power of LLMs(large language models), generating similar triplets to augment existing datasets, minning the potential information in bigger datasets.
SPRC \cite{bai2023sentence} as a recent work, first introduced Qformer which is the new feature of BLIP2 as a fusion encoder, bringing a better underlying network to this community. Sentence-level prompt from SPRC also has enlightening significance.
However, existing methods all focus on the explicit query-target image relation to design their objectives or architectures, the implicit relation, (reference \& target image)-complementary text, is neglected among existing works. In this work, we will jointly utilize the implicit and explicit relation to address the task of composed image retrieval. }


\section{Methodology}
\noindent\textbf{System Overview.}
Figure~\ref{flowchart} depicts the CaLa architecture, characterized by a multimodel branch and an image branch. The multimodal branch serves as a feature extractor for the query pair, incorporating an image encoder and a text encoder for image and text feature extraction, respectively. A cross-encoder is used to produce the multimodal representation of the query pair. The image branch is responsible for encoding the target image.  CaLa begins with feature extraction of the triplets, followed by the imposition of constraints from complementary associations and the primary relation of query-target. As shown in Figure~\ref{flowchart}, the triple features are inputted into a hinge-based cross-attention (HCA) layer to query the target image features and align with the reference image, thereby enforcing the text-bridged image alignment. To impose the constraint of complementary text reasoning, the reference image and target image first pass through a twin attention-based compositor (TAC) to form a combined visual representation, which is then matched with the accompanying complementary text feature. The explicit relation between the query pair and the target image is also incorporated through contrastive learning between the query representation and the target image. The entire network is optimized using the alignment losses derived from the above three alignment procedures.

\noindent\textbf{Preparation.} Assume $I_r$, $C$, $I_t$ to be the reference image, complementary text, and the matched target image, respectively, their features are first encoded using the transformer-based models~\cite{vaswani2017attention,vit}. In detail, the reference image and target image are both encoded using the ViT models~\cite{vit}, but with separate sets of parameters. Following ViTs~\cite{vit,swin}, the image is first evenly split into patches and then fed into the vision transformer for feature extraction. We denote the resulting features of the reference image and target image as $F_r, F_t\in \mathcal{R}^{N\times d}$, respectively, where $N$ is the number of image patches, $d$ is the feature dimension.  For the text, we employ the Bert model~\cite{devlin2018bert} to extract its representation $F_c\in \mathcal{R}^{L\times d}$, where $L$ indicates the length of the text.  With these features, we next elaborate on how to introduce the two complementary associations in subsection~\ref{tia_sec} and subsection~\ref{ctr_sec}. Subsection ~\ref{training_sec} presents the joint network optimization and inference strategy.


\subsection{Text-bridged Image Alignment}
\label{tia_sec}
\vspace{0.2cm}
Text-bridged image alignment (TBIA) aims to integrate the association of the reference image and target images can be aligned with the reference text as the bridge, as shown in Figure~\ref{motivation} (b). To accomplish this, a hinge-based cross-attention (HCA) is designed to query the target image from the reference image with the complementary text as a hidden variable. Subsequently, the attentive feature of the target image and reference image are aligned with a text-guided image alignment constraint.

\begin{figure}[t] 
\centering 
\includegraphics[width=0.28 \textwidth]{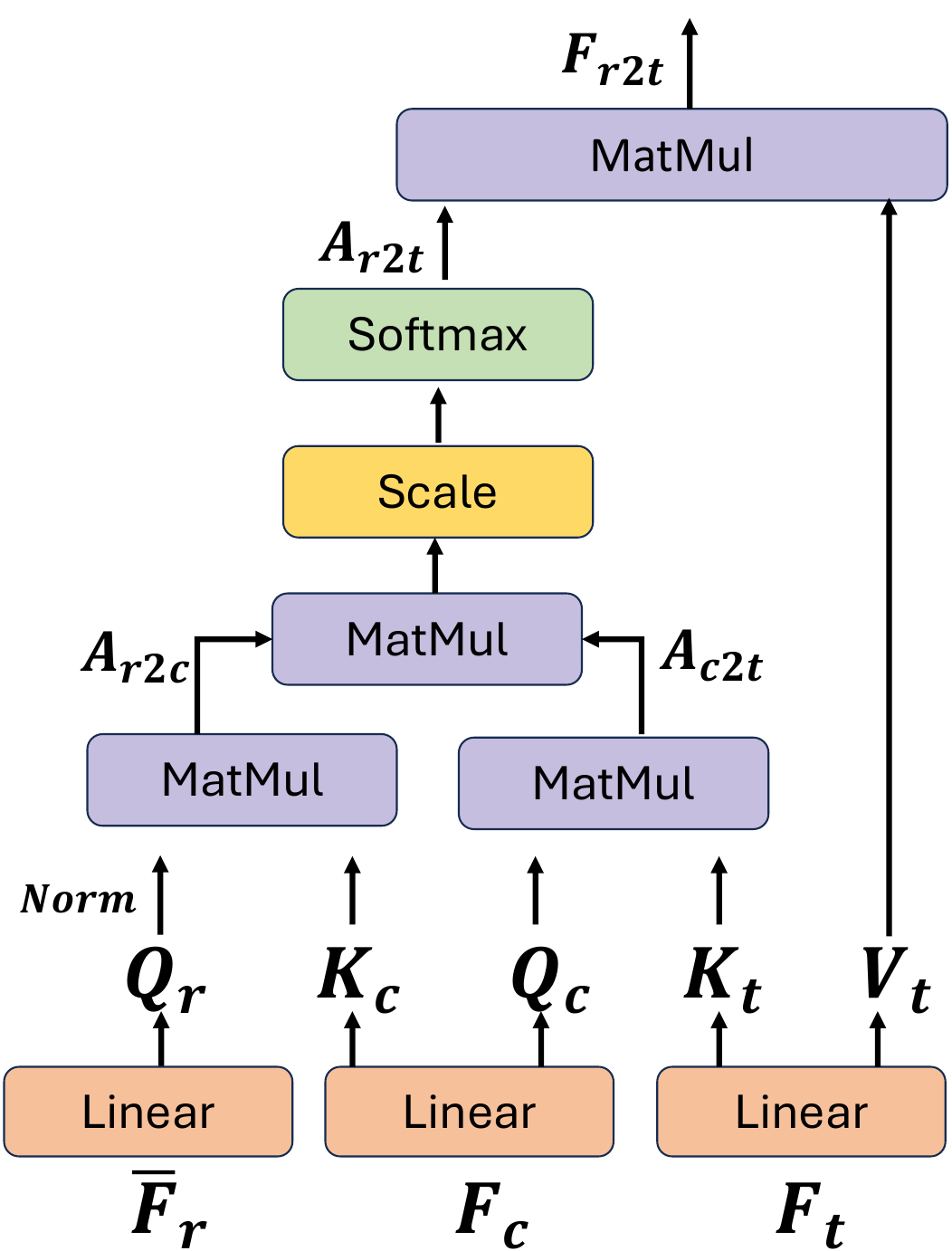} 
\vspace{-0.2cm}
\caption{
The illustration of our hinge-based cross attention. The output of this module can be viewed as a query result from the reference image to the target image, which can be used in the alignment to the reference image.
}
\label{hca} 
\end{figure}

\vspace{0.1cm}
\noindent{\textbf{Hinge-based Cross Attention.}}
As shown in Figure~\ref{hca}, HCA first let the patches of reference image attend to the words in complementary text:
\begin{align}
\label{r2c}
      & Q_r = W_r \times \bar{F}_r,\quad K_c = W_{c} \times F_c, \\
      & A_{r2c}(i,j) = \frac{Q_r(i) \times K_c(j)^{T}}{||Q_r(i)||_2||K_c(j)||_2},
\end{align}
where $A_{r2c}\in \mathcal{R}^{N\times L}$ is the association matrix of reference image querying the text. $W_r$ and $W_c$ are the projection weights of the reference image and the text, respectively. $Q_r(i)$ is the $i$-th feature in $Q_r$, $||\cdot||_2$ means the $l_2$ norm. $\bar{F}_r$ is obtained by passing $F_r$ through the cross encoder with the complementary text $F_c$ as query.  In HAC, we use the attentive $\bar{F}_r$ rather than the straightforward $F_r$ due to two considerations. First, $\bar{F}_r$ can backward the gradient from the TBIA to the cross encoder, thereby benefiting the learning of the cross encoder. Second, querying the attentive features of $F_r$ using the text feature can bridge the semantic gap to aid the subsequent alignment, this will be discussed in our experiments. 

In analogy with the above, we next compute the association with complementary text as the query to attend to the target image:
\begin{align}
\label{c2t}
      & Q_c = W_{c}^{'} \times F_c,\quad K_t = W_t \times F_t, \\
      & A_{c2t}(i,j) = \frac{Q_c(i) \times K_t(j)^{T}}{||Q_c(i)||_2||K_t(j)||_2},
\end{align}
where $A_{c2t}\in \mathcal{R}^{L\times N}$. With the associations of reference-to-text and text-to-target, the association of reference image-to-target image can be formulated as:
\begin{equation}
\label{r2t_attn}
    A_{r2t} = \text{Softmax}(\frac{A_{r2c} \times A_{c2t}}{\sqrt{d}}).
\end{equation}
Eq.~\ref{r2c}-\ref{r2t_attn} employ the complementary text as the hidden point and allow the reference image to attend to the target image, similar to a graph node that the head node links to the tail node through the relation node.

\noindent{\textbf{Text-Guided Image Alignment.}} With the attention matrix of the reference image to the target image, we can directly query the features of the target image: 
\begin{equation}
    F_{r2t} = A_{r2t} \times V_t,\quad \text{where}\quad V_t = W_v\times F_t. 
\end{equation}

In batch data $B$, given the reference image, the matching probability of the pair to the target image via the complementary text is estimated as follows:
\begin{equation}
\label{r2t_loss}
\begin{array}{lll}
    \mathcal{P}(I_t|I_r,C) = \frac{\exp \{\text{sim}(\bar{F}_r,F_{r2t})/\tau\}}{\sum_{I_{t'}\in B}\exp \{\text{sim}(\bar{F}_r,F_{r2t'})/\tau\}}
\end{array}
\end{equation}
where $I_t$ is the target image matched with the query pair $(I_r, c)$, $I_{t'}$ is the unmatched ones, $\tau$ is a temperature parameter and  
 $\text{sim}(\bar{F}_{r},F_{r2t})=\frac{<\bar{F}_{r}, F_{r2t}>}{||\bar{F}_{r}||_2\cdot ||F_{r2t}||_2}$
 is the cosine similarity. Following \cite{radford2021learning}, we set the $\tau$ parameter to 0.1 to ensure that the logits have a sufficient dynamic range in order not to penalize the training process.

Given the above, the overall constraints from text-bridge image alignment reads:
\begin{equation}
    \mathcal{L}_{TBIA} =  -\frac{1}{|B|}
     \sum_{{(I_r, I_t, C)}\in B} \log \mathcal{P}(I_t|I_r,C),   
\end{equation}

\subsection{Complementary Text Reasoning}
\label{ctr_sec}
Complementary text reasoning seeks to model the truth that combining two images can infer the complementary text, as shown in Figure~\ref{motivation} (c). To this end, we first fuse the images into a unified representation via a twin attention-based vision compositor and then align it with the complementary text. 

\vspace{0.1cm}
\noindent\textbf{Twin Attention-based Vision Compositor} is an attention-based architecture~\cite{c:22} designed to effectively fuse images. To achieve this, we employ a twin attention module as the vision compositor, where the reference image and the target image serve as the query and attend to each other, respectively. For the sake of brevity, we will illustrate the procedure using the case of the reference image as the query.
To aid the visual fusion, we feed the reference image forward the image branch, thereby ensuring the reference image lies in the same data flow as the target image, the result feature is noted as $F'_r$\footnote{The re-encoding of the reference image is only employed during training, no extra computational burden is introduced during inference}. 
Given the reference and target image features, the fusion of vision features anchored on the reference image is performed following attention of $M$ layers:
\begin{align}
    H_0^t &= F_t,  \quad l=0, \label{fuse1}\\
    H_m^t &= \text{Attention}(F'_r,H_{m-1}^t,H_{m-1}^t),m\geq 1,  \label{fuse2}
\end{align}
where $\text{Attention}(Q,K,V)$ is the Scaled Dot-Product Attention~\cite{c:22}, $H_l^t$ is the target image-oriented feature from $m$-th layer.

\begin{figure}[t] 
\centering 
\includegraphics[width=0.44 \textwidth]{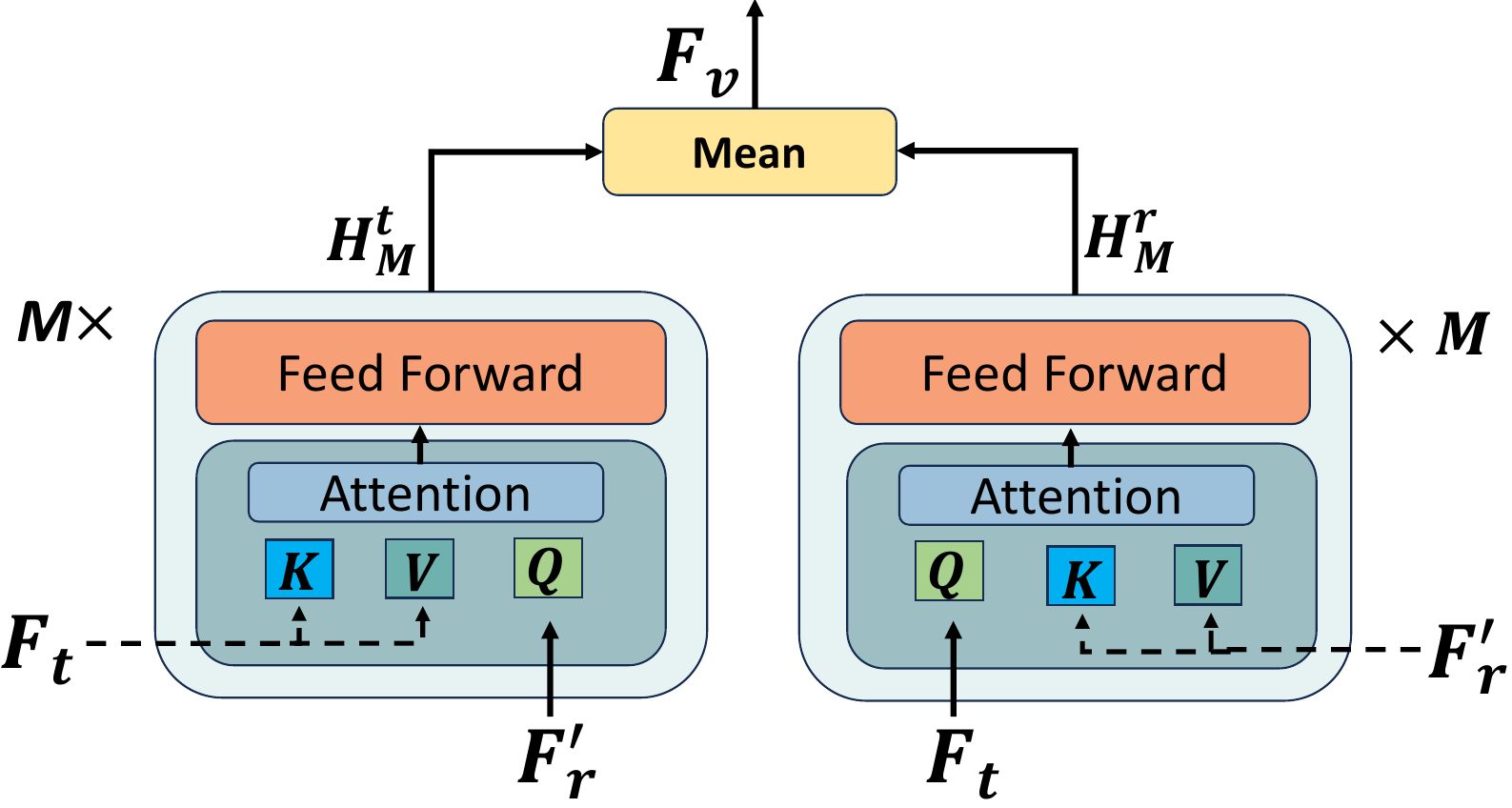} 
\vspace{-0.1cm}
\caption{
The illustration of our Twin Attention-based Vision Compositor. The two branches of cross-attention layers do not share weights and the mean of two CLS tokens is seen as the textual features' counterpart, which is used to reason the complementary text. 
}
\label{hinge-based attetion} 
\end{figure}

During the multiple-layer fusion process, we ensure that the reference image always serves as the query. This approach allows us to accumulate (target image)-oriented visual fusion anchored on the pure reference image, preventing the fusion from being blurred.
We consider the last output of the fusion process as the fused vision feature. Analogously, the fused feature of the (reference image)-oriented fusion, denoted as $H_M^r$, can be obtained by switching the roles of $F'_r$ and $F_t$ in Eq.~\ref{fuse2}. To obtain the final representation of the two images, we average the fused features from the above two views. This gives us the combined visual representation, denoted as $F_v$, calculated as the mean representation: $F_v = (H_M^t+ H_M^r)/2$.

\begin{table*}[t]
\caption{\textbf{CIRR dataset, test set}. $Recall@K$ and $Recall_{subset}@K$(according to \cite{liu2021image}, $Recall_{subset}@1$ best assess fine-grained reasoning ability). Overall 1$^{st}$/2$^{nd}$ in \textbf{bold}/{\bbf{underline}}.$^\dagger$denotes results cited from \cite{baldrati2022conditioned}.}
\label{tab:cirr}
\centering
\begin{tabular}{@{}lcccccccc@{}}
\toprule
\multirow{2}{*}{Method} & \multicolumn{4}{c}{$Recall@K$} & \multicolumn{3}{c}{$Recall_{subset}@K$} & \multirow{2}{*}{$Avg(R@5,R_{sub}@1)$} \\ \cmidrule(lr){2-5} \cmidrule(lr){6-8}
& $K=1$ & $K=5$ & $K=10$ & $K=50$ & $K=1$ & $K=2$ & $K=3$ & \\
\midrule

~\textbf{TIRG}$^\dagger$  & 14.16 & 48.37 & 64.08 & 90.03 & 22.67 & 44.97 & 65.14 & 35.52 \\
~\textbf{TIRG+LastConv}$^\dagger$  & 11.04 & 35.68 & 51.27 & 83.29 & 23.82 & 45.65 & 64.55 & 29.26 \\
~\textbf{MAAF}$^\dagger$  & 10.31 & 33.03 & 48.30 & 80.06 & 21.05 & 41.81 & 61.60 & 27.04 \\
~\textbf{MAAF+BERT}$^\dagger$   & 10.12 & 33.10 & 48.01 & 80.57 & 22.04 & 42.41 & 62.14 & 27.57\\

~\textbf{CIRPLANT}$^\dagger$ & 15.18 & 43.36 & 60.48 & 87.64 & 33.81 & 56.99 & 75.40 & 38.59 \\
~\textbf{CIRPLANT w/OSCAR}$^\dagger$ & 19.55 & 52.55 & 68.39 & 92.38 & 39.20 & 63.03 & 79.49 & 45.88 \\
~\textbf{CLIP4Cir(RN50x4)} & 38.53 & 69.98 & 81.86 & 95.93 & 68.19 &  85.64 & 94.17 &
69.09\\
~\textbf{BLIP2-Cir} & 46.77 & 79.16 & 88.05 & 97.49 & 74.15 & 89.64 & 95.61 &
76.65\\
~\textbf{ARTEMIS(BLIP2)}& 46.46 & 79.13 & \bbf{88.29} &\bbf{97.76} & 74.87 & 90.10 & 96.12 & 77.00 \\
			
\midrule    
~\textbf{$\text{CaLa}_\text{CLIP4Cir(RN50x4)}$}& 35.37 & 68.89 &
80.07 & 95.86 &
66.68 &  84.65 &
93.42 & 67.78 
\\
~\textbf{$\text{CaLa}_\text{ARTEMIS}$}&
\bbf{47.37} & \bbf{79.33} &
88.17 & 97.71 &
\bbf{76.02} & \bbf{90.29} &
\bbf{96.19} & \bbf{77.67} 
\\
~\textbf{$\text{CaLa}_\text{BLIP2Cir}$} & \textbf{49.11} & \textbf{81.21} & \textbf{89.59} & \textbf{98.00} & \textbf{76.27} & \textbf{91.04} & \textbf{96.46} & \textbf{78.74}\\
\bottomrule
\end{tabular}
\end{table*}

\begin{table*}[t]
\caption{\textbf{Fashion IQ, validation set}. We report the challenge metric (\textbf{CM}) and individual $R@K$ scores, where CM=$\frac{R@10+R@50}{2}$
Overall 1$^{st}$/2$^{nd}$ in \textbf{bold}/\bbf{underline}.}
\label{tab:fiqtest}

\centering
\begin{tabular}{@{}llllllllll@{}}
\toprule
\multirow{2}{*}{Method} &  \multirow{2}{*}{\textbf{CM}} & \multicolumn{4}{c}{$R@10$} & \multicolumn{4}{c}{$R@50$} \\ \cmidrule(lr){3-6} \cmidrule(lr){7-10}
 & & Dress & Shirt & Toptee & Mean & Dress & Shirt & Toptee & Mean \\

\midrule
~\textbf{CIRPLANT w/OSCAR} &  
30.20 & 17.45 & 17.53 & 21.64 & 18.87 & 40.41 & 38.81 & 45.38 & 41.53  \\

~\textbf{RTIC-GCN w/GloVe }&  
40.64 & 29.15 & 23.79 & 31.61 & 28.18 & 54.04 & 47.25 & 57.98 & 53.09  \\
~\textbf{CoSMo} &  
39.45 & 25.64 & 24.90 & 29.21 & 26.58 & 50.30 & 49.18 & 57.46 & 52.31  \\
~\textbf{AACL }&  
41.19 &  29.89 & 24.82 & 30.88 & 28.53 & 55.85 & 48.85 & 56.85 & 53.85  \\
~\textbf{DCNet }&  
40.84 &  28.95 & 23.95 & 30.44 & 27.78 & 56.07 & 47.30 & 58.29 & 53.89  \\
~\textbf{SAC w/BERT }&  
41.89 &  26.52 & 28.02 & 32.70 & 29.08 & 51.01 & 51.86 & 61.23 & 54.70  \\
~\textbf{CLIP4Cir(RN50x4) }&  
50.03 & 33.81 & 39.99 & 41.41 & 38.32 & 59.40 & 60.45 & 65.37 & 61.74  \\
~\textbf{ARTEMIS(BLIP2)}&  
56.99 & 40.56 & 46.45 & 49.28 & 45.43 & 65.33 & 67.44 & 72.88  & 68.55  \\
~\textbf{BLIP2-Cir}&  
57.02 & \bbf{41.57} & \bf{46.86} & 49.44 & \bbf{45.96} & 66.02 & 66.00 & 72.25 & 68.09  \\

\midrule

~\textbf{$\text{CaLa}_\text{CLIP4Cir(RN50x4)}$}&  
48.68 & 32.96 & 39.20 & 39.16 & 37.10 & 56.82 & 60.13 & 63.83 & 60.26  
\\

~\textbf{$\text{CaLa}_\text{ARTEMIS}$}&  
\bbf{57.52} & \bbf{40.13} & \bf{46.86} & \bbf{49.87} & 45.62 & \bf{66.88} & \bbf{67.28} & \bf{74.11} & \bbf{69.42}  
\\

~\textbf{$\text{CaLa}_\text{BLIP2Cir}$}&  
\textbf{57.96} & \textbf{42.38} & \bbf{46.76} & \textbf{50.93} & \textbf{46.69} & \bbf{66.08} & \textbf{68.16} & \bbf{73.42} & \textbf{69.22}  \\
\bottomrule
\end{tabular}
\label{fiq}
\end{table*}
									
\vspace{0.1cm}
\noindent{\textbf{Composite Image-Text Matching.}}  In a mini-batch, given the reference and target images, the matching probability of the complementary text is estimated as follows:
\begin{equation}
\label{qtm_loss}
\begin{array}{lll}
    \mathcal{P}(C|I_r, I_t) = \frac{\exp \{\text{sim}(F_c,F_{v})/\tau\}}{\sum_{c'\in B}\exp \{\text{sim}(F_{c'},F_v)/\tau\}},
\end{array}
\end{equation}
where $c'$ is the unaligned text with the two images. Subsequently, this alignment is integrated into the learning of network learning using the following objective. 

\begin{equation}
\label{ctloss}
     \mathcal{L}_{CTR} = -\frac{1}{|B|}
     \sum_{{(I_r, I_t, C)}\in B}\log \mathcal{P}(C|I_r, I_t).
\end{equation}

\subsection{Training and Inference}
\label{training_sec}
\noindent{\textbf{Query-Target Matching}} is a direct and vital relation from the triplet annotation procedures, we also introduce this explicit relation into the network learning. In light of this, we first encoder the multimodal query into one representation, and then align it with the target image. Particularly, we can use Q-former~\cite{blip2} to fuse the embeddings of complementary text and the features of the reference image:
\begin{equation}
    F_q = \text{Q-former}(\text{concat}([Q, F_c], F_r),
\end{equation}
where $Q$ is the learnable prompts in Q-former~\cite{blip2}, $\text{concat}$ means the concatenation operation. As the $F_q$ mainly lies in the vision domain due to the value feature in Q-former's attention derived from the reference image, to implant more guidance of complementary text, we simply add the text feature to $F_q$. Consequently, the feature of the query pair is updated as: $ F_q = \text{Q-former}(\text{concat}([Q, F_c], F_r) + F_c$, which is then alignment with the target image:
\begin{equation}
    \mathcal{L}_{\text{QTM}} = -\frac{1}{|B|}
     \sum_{{(I_r, I_t, C)}\in B}\log \frac{\exp \{\text{sim}(F_q,F_t)/\tau\}}{\sum_{t'\in B}\exp \{\text{sim}(F_{{q}},F_{t'})/\tau\}}.
\end{equation}

\vspace{0.1cm}
\noindent{\textbf{Training Objectives.}} As shown in Figure~\ref{flowchart}, we only optimize part of the network parameters, the two image encoders, the one in the multimodal branch for the reference image and the target image encoder, are frozen.  The parameters are optimized using the presented three constraints:  
\begin{equation}
    \mathcal{L} = \mathcal{L}_{\text{QTM}} (\Theta_{t},\Theta_{v})+ \alpha\mathcal{L}_{\text{TBIA}} + \beta\mathcal{L}_{\text{CTR}},
\end{equation}
where $\alpha$ and $\beta$ are two trade-off hyper-parameters.

\vspace{0.1cm}
\noindent{\textbf{Inference}.} Note that the proposed two complementary associations are employed during training as the auxiliary objectives, such that no extra computation or parameter burden is introduced during inference. For a query pair and a gallery of candidate target images, the ranking scores are computed using the multimodal representation and image feature presented in query-target matching.

\section{Experiment}
\subsection{Implementation Details.}
The disclosed complementary associations are widely beneficial relations for composed image retrieval. To support this claim, we equip the proposed CaLa on the top of three baseline methods, including CLIP4Cir~\cite{baldrati2022conditioned}, ARTEMIS~\cite{delmas2022artemis}, and BLIP2-Cir~\footnote{BLIP2-Cir is an improved version of CLIP4Cir by replacing the backbone with BLIP2.}. CLIP4Cir and BLIP2-Cir are of similar architectures but different feature extractors, while the architecture of ARTEMIS differs from the two above-mentioned significantly.  With these baselines, we can make a comprehensive evaluation of our CaLa under different backbones and architectures. Due to the page limitation, we give the training details using BLIP2Cir as the backbone, the cases of  CLIP4Cir and ARTEMIS are familiar.
\footnote{
Note that CLIP4Cir is a two-stage training process that can provide two backbones, CLIP4Cir-Sum and CLIP4Cir-Combiner. Here we take CLIP4Cir-Sum as the backbone which means, only the text encoder is fine-tuned.
For ARTEMIS, we adopt its fusion architecture and substitute its encoder with BLIP2. This replacement diminishes the influence of backbones, enabling experiments to assess the suitability of CaLa across various architectures.
}

The model is trained with Pytorch on 1 NVIDIA 4090 GPU. We train the network with a mini-batch of 64 for 30 epochs. We follow the design of BLIP-2, the visual, textual, and multi-modal encoders are initialized from the BLIP-2 pretrained model with ViT-L. We resize the input image size to 224 × 224 and with a padding ratio of 1.25 for uniformity. The learning rate is initialized to 5e-6 and 1e-5 following a cosine schedule for the CIRR and Fashion-IQ datasets, respectively. The twin attention-base compositor is implemented with two multi-head attention modules.  In TAC, attention weights within the same branch are shared, while weights between the two branches are independent. The number of attention layers is set to 4 in TAC, referring to the best in Table \ref{tab:ablation:tac_layers}. Hyper-parameters $\alpha$ and $\beta$ are fixed to 0.45 and 0.1, respectively, referring to Table \ref{tab:ablation:alpha} and Table \ref{tab:ablation:beta}. \\

\subsection{Datasets and Evaluation Metrics.} 
We employ the widely-used  CIR benchmarks \textbf{CIRR}~\cite{liu2021image} and \textbf{FahsionIQ}~\cite{fashionIQ} for performance evaluation.
CIRR comprises 21,552 real-life images sourced from the NLVR2 dataset~\cite{nlvr2}.
We report Recall at  1,5,10 and 50 ranks for performance comparison. The CIRR dataset also includes a subset for fine-grained differentiation. This subset consists of negative images that exhibit high visual similarity. In line with established practices~\cite{delmas2022artemis,baldrati2022conditioned}, we adhere to this and report the $\text{Recall}_\text{subset}$ metrics at ranks 1, 2 and 3.

FashionIQ is a dataset specifically designed for fashion-conditioned image retrieval. It comprises 30,134 triplets generated from a collection of 77,684 web-crawled images. The dataset is categorized into three distinct fashion categories: Dress, Toptee, and Shirt. Following the previous works~\cite{lee2021cosmo,delmas2022artemis,baldrati2022conditioned}, we report the Recall at 10 and 50 ranks of the three categories to evaluate the performance.

To ensure the robustness of our method, we opted for one open-domain dataset and one fashion-domain dataset, aligning with the approach taken by CLIP4Cir \cite{baldrati2021conditioned}. This selection has been consistently adopted by many subsequent methods~\cite{liu2024bi,liu2023candidate}. Notably, we did not utilize the Fashion200K ~\cite{han2017automatic} and Shoes ~\cite{guo2018dialog} datasets, as FashionIQ is a relatively new and representative dataset.

In terms of choosing evaluation metrics, considering that both mentioned datasets contain only one positive sample for each query, Recall at K suffices for assessing the method's accuracy, which is why we opted for Recall at K over Mean Average Precision (MAP) as our evaluation metric.

\begin{figure*}[t] 
\centering 
\includegraphics[width=0.8\textwidth]{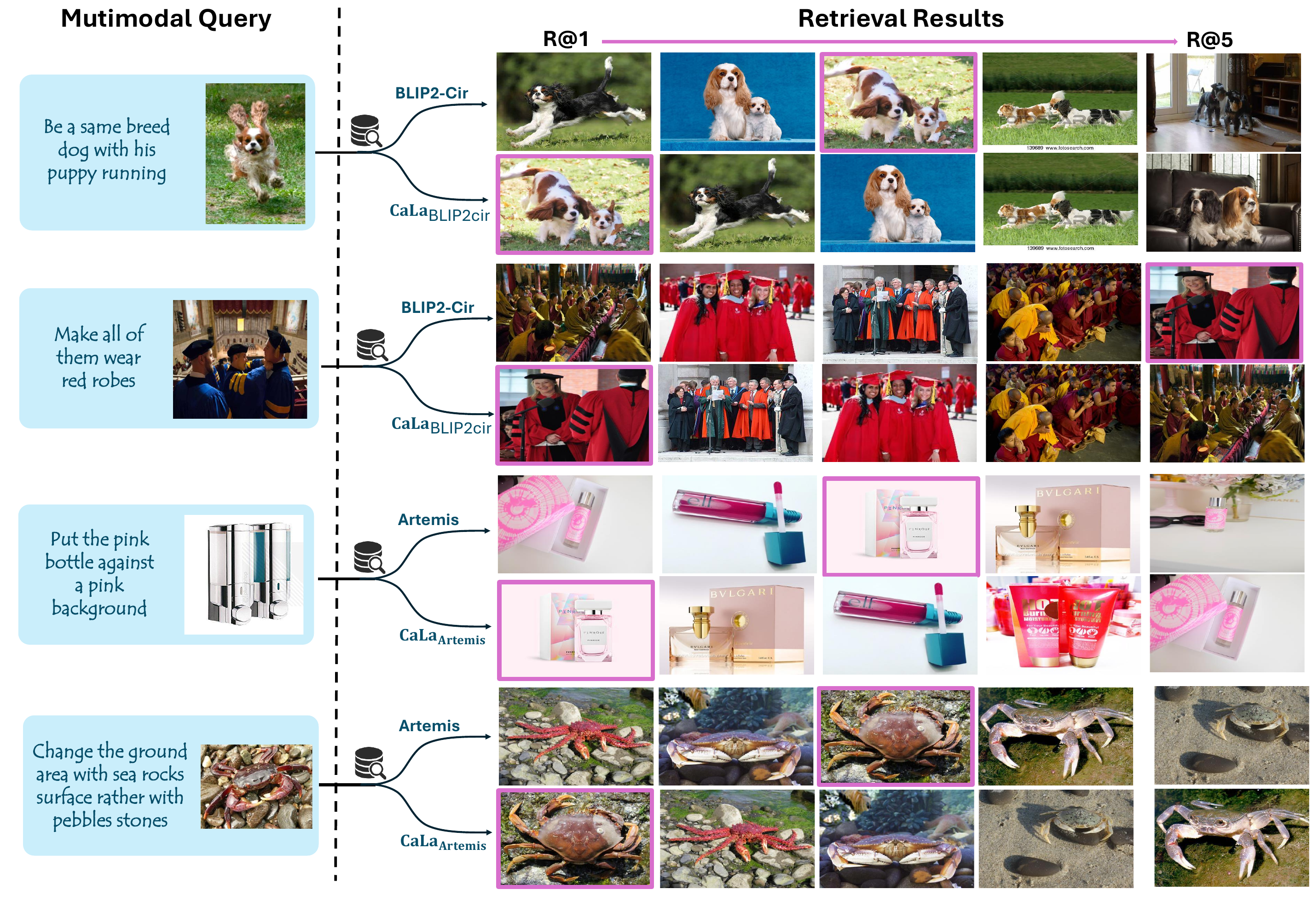}
\vspace{-0.4cm}
\caption{Qualitative results on CIRR validation dataset. We show the results of both baseline solely and with CaLa for a clear comparison: BLIP2Cir \emph{vs} $\text{CaLa}_\text{BLIP2Cir}$, and ARTEMIS \emph{vs} $\text{CaLa}_\text{ARTEMIS}$. Images in red boxes are the target images responding to the query pair. We can find that the target image can be identified more accurately when our CaLa is equipped.} 
\label{visualization} 
\end{figure*}

\begin{figure}[t] 
\centering 
\includegraphics[width=0.44 \textwidth]{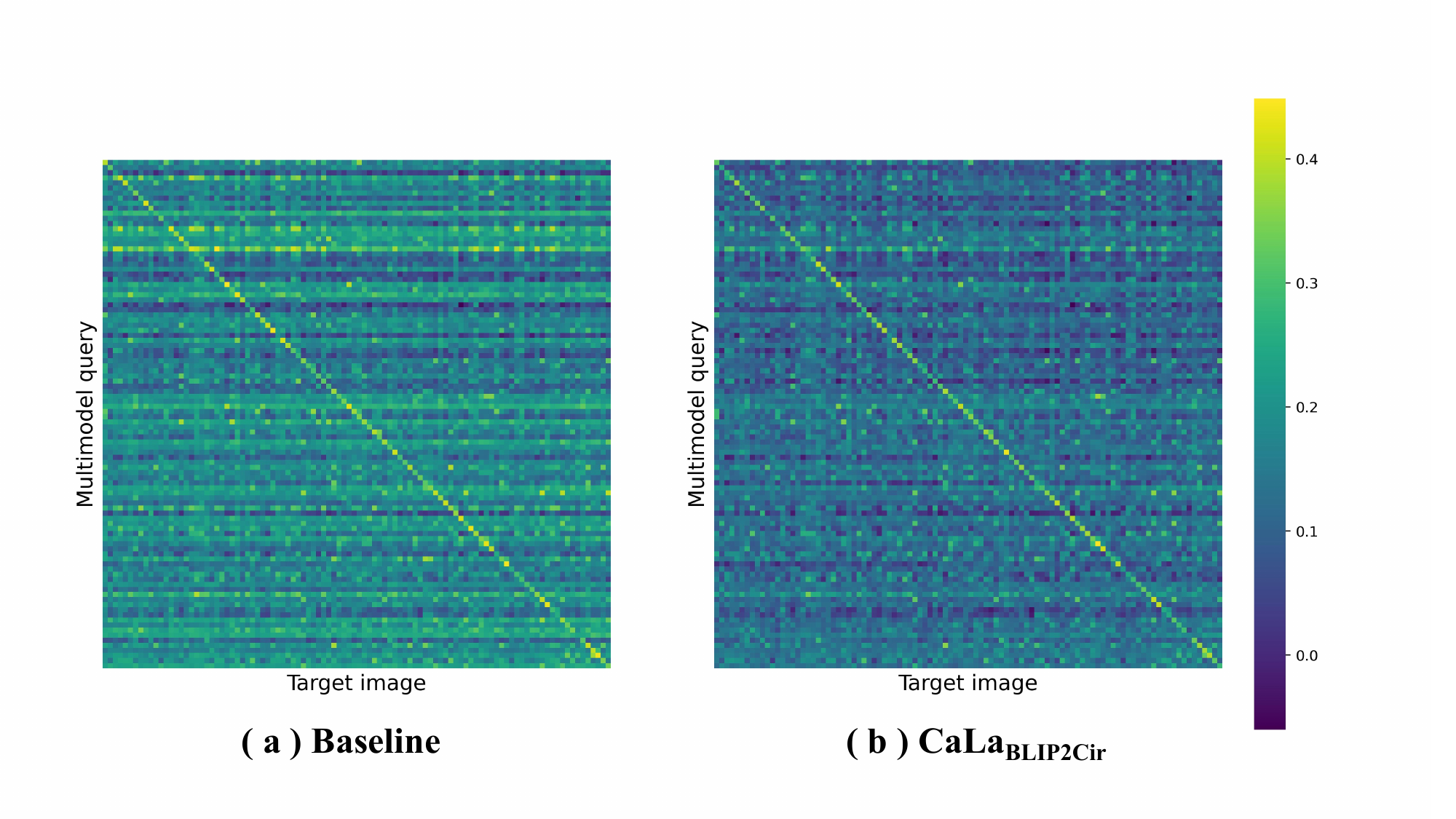} 
\vspace{-0.3cm}
\caption{
Visualization of the similarity matrix of multi-
modal query and target image.
}
\label{heapmap} 
\end{figure}


\subsection{Quantitative Results}
Table~\ref{tab:cirr} and Table~\ref{tab:fiqtest} report the performance with the state-of-the-art methods~\cite{liu2021image,shin2021rtic,lee2021cosmo,kim2021dual,delmas2022artemis,jandial2022sac,baldrati2022conditioned} on CIRR and FashionIQ dataset, respectively, from which we can observe that the models with our CaLa can surpass the base methods by a clear margin. $\text{CaLa}_\text{BLIP2Cir}$ means the BLIP2Cir serve as the baseline model.
Table~\ref{tab:cirr} shows the comparison between our method and current state-of-the-art models on the CIRR test set. These quantitative results are obtained through the official evaluation server.
As shown in Table~\ref{tab:cirr}, $\text{CaLa}_\text{BLIP2Cir}$ achieves a notable average recall of 78.74\% on the CIRR dataset. In contrast, the base method, BLIP2-Cir, achieves an average recall of 76.65\%. Additionally, our CaLa outperforms all other competitive methods across various ranks, ranging from 1 to 50. Particularly in low-rank recall measures, we observe an improvement of 2.09\% on R@5 compared to the base method. Notably, within the fine-grained differentiation subset, our method demonstrates even better performance with a substantial improvement of 2.12\% on $\text{Recall}_\text{subset}$@1. This improvement indicates the effectiveness of our method in handling fine-grained retrieval tasks. When CLIP4Cir and ARTEMIS serve as the backbones, our CaLa can also promote the performance, revealing the disclosed associations in CaLa are widely beneficial for composed image retrieval.\\ 

The performance on FashionIQ shown in Table ~\ref{fiq} is also conclusive,  our CaLa can attain better performance than all comparison approaches, where the average  Recall@50 on three categories can reach 69.22\%, surpassing the baseline by 1.13\%.

\begin{figure*}[t] 
\centering 
\includegraphics[width=\textwidth]{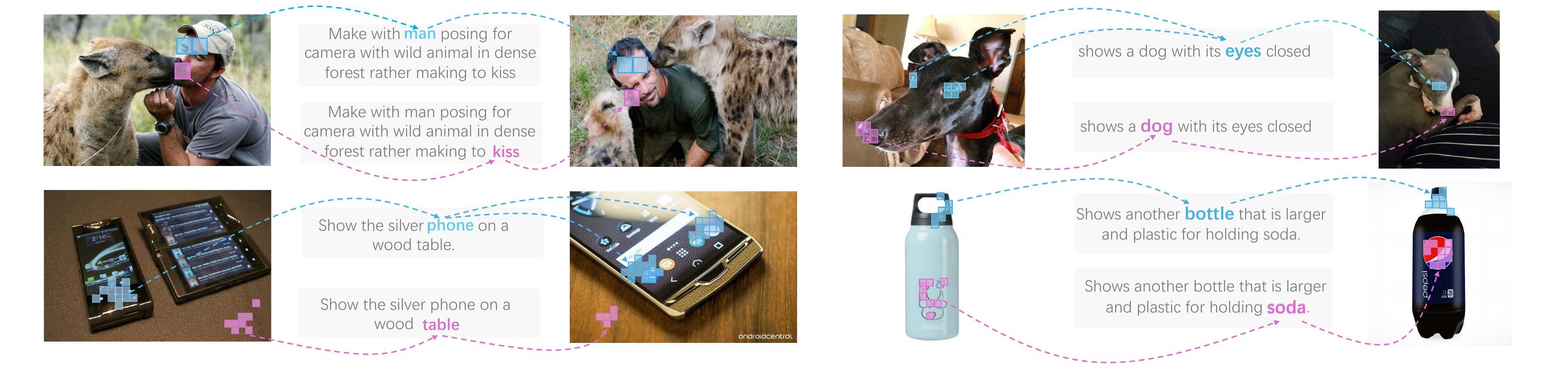}
\vspace{-0.3cm}
\caption{\textbf{Attention visualization in HCA.} (Produced by $\text{CaLa}_\text{BLIP2Cir}$), reference image, complementary text, and target image are placed from left to right in sequence. In both attention procedures, namely the reference image attending to text and text attending to the target image, the essential components (patches or words) effectively attend to the corresponding critical components in their counterpart. (The patch size is varied since the image is scaled for a neat exhibition.)  
} 
\label{hcattn} 
\end{figure*}

\begin{table*}[t]
\caption{\textbf{Ablation with different backbones on CIRR dataset, val set}. $Recall@K$ and $Recall_{subset}@K$(according to \cite{liu2021image}, $Recall_{subset}@1$ best assess fine-grained reasoning ability). 
}
\label{cirr}
\centering
\begin{tabular}{@{}lcccccccccc@{}}
\toprule
\multirow{2}{*}{Backbone} & & &\multicolumn{4}{c}{$Recall@K$} & \multicolumn{3}{c}{$Recall_{subset}@K$} & \multirow{2}{*}{$Avg(R@5,R_{sub}@1)$} \\ 
\cmidrule(lr){4-7} \cmidrule(lr){8-10}
& TBIA & CTR & $K=1$ & $K=5$ & $K=10$ & $K=50$ & $K=1$ & $K=2$ & $K=3$ & \\
\midrule
\multirow{4}*{BLIP2Cir} &
- & -  & 46.78 & 78.88 & 87.75 &	97.54 &	74.55 &	89.24 &	95.55 & 76.72 \\
 &		
\checkmark & - & 48.03 &	80.15 &	88.47 &	97.46 &	76.29 &	90.89 &	95.96 &	78.22 \\
 &		
- & \checkmark & 47.81 &	80.67 &	89.21 &	97.68 &	75.82 &	90.15 &	95.67 &	78.25 \\
 &
\checkmark & \checkmark & 48.86 &	80.48 &	88.47 &	97.44 &	76.37 &	90.31 &	95.84 &	78.43 \\
\midrule
							
\multirow{4}*{Artemis} &
- & -  & 45.99  &78.67  &87.32  &96.91 &74.46	 &89.36	 &95.48	 &76.56  \\
 &									
\checkmark & - &45.66  &78.86	 &87.66	 &97.13	 &74.84	 &89.62	 &95.72	 &76.84	 \\
 &									
- & \checkmark &47.26  &79.07	 &87.95	&97.44 &75.82	 &90.39	 &95.74	 &77.45 	 \\
 &							
\checkmark & \checkmark &47.81  &79.65 &87.97 &97.37 &75.56 &90.39 &96.03 &77.60	 \\
\midrule

\multirow{4}*{CLIP4Cir} &							
- & -  &32.62  &67.02  &79.74  &95.31 &65.41 &84.67 &92.54 &66.22 \\
 &		
				
\checkmark & -  &34.68  &69.58	 &82.01 &96.14 &67.40 &86.08 &93.92 &68.49 \\
 &									
- & \checkmark  &35.69  &69.12	 &80.79	&95.58 &70.58	 & 87.78 &94.07 & 69.85\\
 &
\checkmark & \checkmark &34.25  &69.98 &81.92 &96.13 &67.28 &86.18 &93.73 &68.63 \\

\bottomrule
\end{tabular}
\end{table*}

\subsection{Qualitative Results}
\textbf{Retrieval Results.} Referring to retrieval results from Figure~\ref{visualization}, it shows that our method can better cope with some queries, pushing irrelevant images away from the low-rank results.
In order to figure out the reason for this, we obtain the multi-modal query features through both BLIP2-Cir solely and with CaLa, and then compute the cosine similarity between them and the target features extracted from BLIP2 visual encoders. From Figure~\ref{heapmap}, we can see that our CaLa facilitates the alignment of the positive pairs, in the meantime, diluting the hard negative pairs' similarities globally. 
We attribute this to the utilization of two complementary associations, giving more comprehensive guidance for network learning such that the learned model can identify the target image more accurately.

\vspace{0.1cm}
\noindent\textbf{Attention Visualization in HCA.} Figure~\ref{hcattn} presents a visual representation of the four instances of attention in HCA, which demonstrates how HCA perceives the relationships within the triplets. It can be observed that the critical patch (word) effectively attends to its semantically aligned word (patch). For example, in the top-left example, the ``kiss'' patches in the reference image (left-most) attend to the word ``kiss'', which is aligned with the ``kiss'' patches in the target image (right-most) through the attention mechanism.  This observation confirms that HCA functions as intended.

\subsection{Ablation Study}
In this section, we take $\text{Cala}_\text{BLIP2Cir}$ as the base method and conduct experiments on the CIRR validation set to verify the key components in our CaLa, and discuss the configurations of our designs including the architecture and hyper-parameters to make everything clear. 

\vspace{0.1cm}
\noindent\textbf{Efficacy of TBIA and CTR.} 
To access the respective contributions of TBIA and CTR, 
we conduct experiments on CIRR and challenge the different module choices by training them independently,  the results are reported in Table~\ref{cirr}.
We can observe clear performance improvements by incorporating both TBIA and CTR into our baseline methods. The inclusion of TBIA results in recall enhancements of 1.27\% for Recall@5, and 1.74\% for $\text{Recall}_\text{subset}$@1 on the CIRR validation set. When applying CTR to the baselines, we also observe the expected performance improvements of 1.79\% for Recall@5 and 1.27\% for $\text{Recall}_\text{subset}$@1. By simultaneously including both TBIA and CIR, we further advance the performance and achieve our best results. These observations reveal that the complementary associations uncovered by this work can not only independently improve Composed Image Retrieval (CIR) performance but also assist in the widely-used explicit association learning, well verifying its potential and significance for composed image retrieval.

\begin{table*}[t]
\caption{\textbf{Discussion}. We use $\text{CaLa}_\text{BLIP2Cir}$ as the backbone and discuss the detailed configurations of our HCA and TAC (a)-(d), as well as the hyper-parameters $\alpha$ and $\beta$ (e-f).
}
\centering
\subfloat[
Pure vs Attentive Features of the Reference image in HCA. The attentive features from the text can attain better results.
\label{tab:ablation:pure_attentive}
]{
\centering
\begin{minipage}{0.29\linewidth}{\begin{center}
\tablestyle{2pt}{1.1}
\begin{tabular}[t]{@{}lcccccc@{}}
Ref. Feat. &
 $R@5$ & $R_{sub}@1$ & $Avg(R@5,R_{sub}@1))$ \\
\shline
Baseline &	
78.88 & \bbf{74.55} & 76.72 \\ 
Pure feat. &		
\bbf{79.31} & 74.53 & \bbf{76.92} \\ 
Attn. feat. &		
\bf{80.15} & \bf{76.29} & \bf{78.22} \\ 

\end{tabular}
\end{center}}\end{minipage}
}
\hspace{0.6em}
\subfloat[
Share \emph{vs} None-share Text feature in HCA. Sharing the text feature in the two-stage attention of HCA leads to better results.
\label{tab:ablation:share-key}
]{
\begin{minipage}{0.29\linewidth}{\begin{center}
\tablestyle{4pt}{1.1}
\begin{tabular}[t]{@{}lcccccc@{}}
HCA Arch. &
 $R@5$ & $R_{sub}@1$ & $Avg(R@5,R_{sub}@1))$ \\
\shline
Baseline &	
78.88 & 74.55 & 76.72 \\ 
Share &		
\bbf{80.15} & \bf{76.30} & \bf{78.22} \\ 
Non-share &		
\bf{80.29} & \bbf{75.94} & \bbf{78.12} \\ 
\end{tabular}
\end{center}}\end{minipage}
}
\hspace{2.5em}
\subfloat[
Share \emph{vs} Non-share Weights in TAC. In the case of non-sharing the weights can bring more performance gain. 
\label{tab:ablation:tac_share}
]{
\begin{minipage}{0.29\linewidth}{\begin{center}
\tablestyle{1pt}{1.1}
\begin{tabular}[t]{@{}lcccccc@{}}
Share weight &
 $R@5$ & $R_{sub}@1$ & $Avg(R@5,R_{sub}@1))$ \\
\shline
Baseline &	
78.88 & 74.55 & 76.72 \\ 

Share &		
\bbf{80.53} & \bbf{75.41} & \bbf{77.97} \\ 
		
Non-share &		
\bf{80.67} & \bf{75.82} & \bf{78.25} \\ 
\end{tabular}
\end{center}}\end{minipage}
}
\\
\centering
\vspace{1em}
\subfloat[
Effect of TAC layers. We stack 4 cross-attention layers in TAC can harvest the best performance on average.
\label{tab:ablation:tac_layers}
]{
\begin{minipage}{0.29\linewidth}{\begin{center}
\tablestyle{2pt}{1.1}
\begin{tabular}[t]{@{}lcccccc@{}}
TAC layers &
 $R@5$ & $R_{sub}@1$ & $Avg(R@5,R_{sub}@1))$ \\
\shline
Baseline &	
78.88 & 74.55 & 76.72 \\ 

1x &		
 80.60 &  75.05 & 77.82  \\ 
2x &		
 80.46 &  \bbf{75.34}  &  \bbf{77.90} \\ 
3x &		
 80.24 &  75.20 &  77.72\\ 
    
4x &		
 \bbf{80.67} &\bf{75.82 } & \bf{78.25}  \\ 
5x &		
 \bf{80.70} &  75.00 &  77.85 \\ 

\end{tabular}
\end{center}}\end{minipage}
}
\hspace{2em}
\subfloat[
Effect of hyper-parameter $\alpha$. The case of $\alpha=0.45$ performs better than other cases in terms of the average performance.
\label{tab:ablation:alpha}
]{
\centering
\begin{minipage}{0.29\linewidth}{\begin{center}
\tablestyle{4pt}{1.1}
\begin{tabular}[t]{@{}lcccccc@{}}
$\alpha$ &
 $R@5$ & $R_{sub}@1$ & $Avg(R@5,R_{sub}@1))$ \\
\shline
$\alpha$ = 0 &	
78.88 & 74.55 & 76.72 \\ 

$\alpha$ = 0.40 &		
79.86 & \bbf{75.81} & 77.84 \\ 

$\alpha$ = 0.45 &		
80.67 &  \bf{75.82} & \bf{78.25} \\ 
$\alpha$ = 0.50 &		
\bf{80.91} &  75.25 & \bbf{78.08} \\ 
$\alpha$ = 0.55 &		
\bbf{80.79} &  75.36 & 78.02 \\ 

\end{tabular}
\end{center}}\end{minipage}
}
\hspace{2em}
\subfloat[
Effect of hyper-parameter $\beta$. 0.1 is a good choice$\beta=0.1$, surpassing the other cases by a clear margin. 
\label{tab:ablation:beta}
]{
\begin{minipage}{0.29\linewidth}{\begin{center}
\tablestyle{4pt}{1.1}
\begin{tabular}[t]{@{}lcccccc@{}}
$\beta$ &
 $R@5$ & $R_{sub}@1$ & $Avg(R@5,R_{sub}@1))$ \\
\shline
$\beta$ = 0 &	
78.88 & 74.55 & 76.72 \\ 


$\beta$ = 0.1 &		
\bbf{80.15} & \bf{76.30} & \bf{78.22} \\ 
$\beta$ = 0.2 &		
\bf{80.22} &  \bbf{74.58} & \bbf{77.40} \\ 
$\beta$ = 0.3 &		
\bf{80.22} & \bbf{74.58} & \bbf{77.40} \\ 

\end{tabular}
\end{center}}\end{minipage}
}
\label{tab:ablations}
\vspace{-.5em}
\end{table*}

\vspace{0.1cm}
\noindent\textbf{Pure \emph{vs} Attentive Features of Reference image in HCA.} In HCA, we tested both the attentive and pure reference features to figure out whether or not the text features can serve as a pivot to bridge the semantic gap, aidding the subsequent alignment. From Table \ref{tab:ablation:pure_attentive} we can see pure features barely promote the alignment procedure while attentive features of reference image yield an advancement. This also confirms to some extent our hypothesis that text has the capability to serve as the pivot bridging the two images.

\vspace{0.1cm}
\noindent\textbf{Shared \emph{vs} Non-shared Text Feature in HCA.}
We also discuss whether or not to share the text features in HCA, which refers to the sharing of $W_{c}$ and 
$W_{c}^{'}$ or not. We have developed a set of comparative experiments to validate which is the more appropriate setting.
From Table \ref{tab:ablation:share-key} we can see that sharing the weight performs better. We attribute this to sharing the $W_{c}$ to make the text features work better as a bridging pivot between the two images, further confirming the plausibility of the hinge-based cross attention. 
Hence, in experiments other than this one, we all follow this setting of sharing weights between $W_{c}$ and 
$W_{c}^{'}$.


\vspace{0.1cm}
\noindent\textbf{Share \emph{Vs} Non-share Weights in TAC.} In our Twin Attention-based vision compositor (TAC), the default configuration involves sharing the attention weights for fusing the images. 
To evaluate the impact of shared weights between the two branches, we compare the performance of these two cases, as presented in Table~\ref{tab:ablation:tac_share}. The results demonstrate that non-shared weights lead to better performance compared to the shared weights configuration.
We attribute the performance improvement to utilizing different parameters in the two branches. Using the same parameters for reasoning from the target image to the reference image may produce inconsistent results during alignment.

\vspace{0.1cm}
\noindent\textbf{Effect of Attention Layers in TAC.}  We fix the attention layers in our TAC at 4, 
in this subsection, 
we study the performance change \emph{w.r.t} the attention layers. The performance tendency \emph{vs} layer numbers are shown in Table~\ref{tab:ablation:tac_layers}, we observe that performance increases with the number of layers, peaks at 4 layers, and then declines. Hence, in experiments other than this one, we set the number of layers to 4.
We attribute
the increase in performance of the iterative query procedure,
producing an accumulated (target image)-oriented visual fusion anchored on the pure reference image and vice versa. 

\vspace{0.1cm}
\noindent\textbf{Effect of hyper-parameter $\alpha$ and $\beta$.} We adopt a weighted average strategy when introducing two complementary losses, referring to the hyper-parameters $\alpha$ and $\beta$. We carried out certain experiments in the effect of the two weight hyper-parameters, performance increases with which, peaks at 0.45 and 0.1 separately then declines. Hence, in experiments other than this one, we set $\alpha$ and $\beta$ to 0.45 and 0.1, respectively.

\section{Conclusion}

In this work, we present a new thinking for composed image retrieval and uncover two complementary associations by treating the annotated triplets as graph nodes. Particularly, the reference image, complementary text, and target image are analogous to the head, relation, and tail in a graph node, respectively. As a result, two relations in the triplet, which are both overlooked by previous works, are mined. First, the reference image and target image can be aligned through the complementary text. To integrate this association, we propose a hinge-based cross-attention to query the target image from the reference image via the complementary text, subsequently, the attentive target features are aligned with the reference image. The second found relation is that the complementary text can be inferred by compositing the two images, similar to the link prediction in knowledge graph learning. To equip this association, we propose a twin-attention-based image compositor to fuse the image features, and then the resulting features are aligned with the representation of complementary text. We conducted extensive quantitative and qualitative experiments on two widely-used benchmarks CIRR and FashionIQ, the results show that the contributed two complementary associations are beneficial to composed image retrieval.

\section{ACKNOWLEDGEMENTS}
This work was supported by the NSFC project under grant No. 62302140, in part by the NSFC under grants 62272380 and 62103317.

\bibliographystyle{ACM-Reference-Format}
\bibliography{sample-base}

\end{document}